# Graphical Models and Exponential Families

**Dan Geiger\* and Christopher Meek**
Microsoft Research
Redmond WA, 98052-6399
dang@cs.technion.ac.il   meek@microsoft.com

## Abstract

We provide a classification of graphical models according to their representation as subfamilies of exponential families. Undirected graphical models with no hidden variables are linear exponential families (LEFs), directed acyclic graphical models and chain graphs with no hidden variables, including Bayesian networks with several families of local distributions, are curved exponential families (CEFs) and graphical models with hidden variables are stratified exponential families (SEFs). An SEF is a finite union of CEFs satisfying a *frontier* condition. In addition, we illustrate how one can automatically generate independence and non-independence constraints on the distributions over the observable variables implied by a Bayesian network with hidden variables. The relevance of these results for model selection is examined.

## 1   Introduction

A graphical model is a family of probability distributions. The set of distributions associated with a graphical model is usually defined in one of the following two ways. Implicitly, by a set of constraints that the distributions must satisfy, or, parametrically (explicitly), by a mapping from a set of parameters to a set of distributions. Graphical models without hidden variables can be defined implicitly using a set of independence constraints. Graphical models with hidden variables, on the other hand, are usually defined parametrically because the non-independence constraints on the distributions over the observable variables are not easily established. In this paper, we discuss a procedure for identifying non-independence (and independence) constraints true of the distributions in a parametrically defined family. In addition, we provide a classification of graphical models according to their representation as subfamilies of exponential families. The relevance for model selection of both the procedure and the classification is examined.

The properties of linear exponential families (LEFs) have been extensively studied in statistics. We begin by reviewing LEFs and demonstrating the well-known fact that undirected graphical models with no hidden variables are LEFs (e.g., Lauritzen, 1996). This result provides access to the properties of these families. In particular, with respect to model selection, Schwarz (1978) established, under some regularity assumptions, that the Bayesian Information Criteria (BIC) is a valid asymptotic rule for selecting models from a collection of linear exponential families.

We show that directed acyclic graphical models and chain graphs with no hidden variables are curved exponential families (CEFs). Roughly, a curved exponential family is a set of distributions such that (1) each member of the set is an exponential family distribution and (2) the family corresponds to a smooth manifold. Showing that some graphical models are CEFs provides access to the properties of these families which have been studied in the statistics community (e.g., Kass and Vos, 1997). In particular, Haughton (1988) established, under some regularity assumptions, that BIC is a valid asymptotic rule for selecting models from a set of curved exponential families. Asymptotic techniques are not needed for choosing between multinomial Bayesian networks because a closed-form formula for the marginal likelihood $P(data|model)$ is known for these classes (Cooper and Herskovits, 1992). Such a formula is not known for various Bayesian networks including those using noisy or-gates or causal independence models and so our classification of these graphical models as CEFs justifies using BIC for selecting models within these classes.

We also show that graphical models with hidden variables are stratified exponential families (SEFs). An SEF is a finite union of CEFs satisfying a *frontier* condition. We show that SEFs arise naturally from both implicitly defined and parametrically defined graphical models. With respect to model selection, Geiger, Heckerman, and Meek (1996) note that although re-searchers have been using BIC for selecting models among Bayesian networks with hidden variables, this methodology has not yet been established as an asymptotic approximation of a Bayesian procedure as it has for CEFs.

\* On sabbatical from the Technion, Israel.

Finally, we discuss a procedure called *implicitization* that takes a parametrically defined graphical model and generates both independence and non-independence constraints. We illustrate the implicitization procedure on several Bayesian networks with hidden variables finding both new and previously known constraints on the distributions over the observable variables. Since these constraints vary from one model to another they can be used to distinguish between models. For instance, when we applied implicitization to a particular Gaussian Bayesian network we generated the well known tetrad difference constraints which have been used for model selection and evaluation by several researchers (see, e.g., Spirtes, Glymour, and Scheines, 1993).

## 2   Linear Exponential Families

Exponential families of probability distributions play an important role in the study of statistics. This class is essentially the only class of distributions that have finite dimensional sufficient statistics. In this background section we give a definition of linear exponential families (LEFs) and discuss the well-known representation of undirected graphical models as LEFs (e.g., Lauritzen, 1996).

### 2.1   Definition of Linear Exponential Families

A *family* (or model) is a set of probability density functions. A probability density in an exponential family is given by

$$p(x|\eta) = e^{<\eta, t(x)> - \psi(\eta)} \qquad (1)$$

where $x$ is an element of a sample space $\mathcal{X}$ with a dominating measure $\mu$ and $t(x)$ is a sufficient statistics defined on $\mathcal{X}$ taking values in $R^k$ with an inner product $< .,. >$. The sample space $\mathcal{X}$ is typically either a discrete set, $R^n$, or a product of these. In the later case, in this paper, the product sample space is viewed as a finite set of variables each having a domain which is either finite or $R$. The quantity $\psi(\eta)$ is the normalization constant.

Every probability distribution for a finite sample space $\mathcal{X}$ belongs to an exponential family. For example, a sample space that consists of four outcomes can be written in the form of Eq. (1) by choosing $t(x)$ and $\eta$ as follows: $t(x) = (t_1(x), t_2(x), t_3(x))$ where $t_i(x) = 1$ if $x$ is outcome $i$, $1 \leq i \leq 3$, and zero otherwise, and $\eta_i = \log(w_i/w_0)$ where $w_i$ is the probability of outcome $i$, $1 \leq i \leq 3$, and $w_0 = 1 - \sum_{i=1}^{3} w_i$ is the probability of the forth outcome.

When the vector $\eta$ has $k$ coordinates and when $p(x|\eta)$

cannot be represented with a parameter vector smaller than $k$, then the representation is *minimal* and the *order* (or *dimension*) of this family is $k$, and the parameters are called *natural parameters*. It is known that this order is unique for each family. The natural parameter space is given by

$$N = \{ \eta \in R^k \mid \int e^{t(x)\eta - \psi(\eta)} d\mu(x) < \infty \}$$

The set of probability distributions having the form (1) are denoted by $\mathcal{S}$. If for each $\eta$ in $N$ there exists $P_\eta$ in $\mathcal{S}$, then $\mathcal{S}$ is said to be *full* exponential family; if, in addition, $N$ is an open subset of $R^k$, then $\mathcal{S}$ is said to be a *linear* exponential family. The name linear exponential family draws from the fact that the log densities form a vector space over $R$ were the coordinates of $t(x)$, called the canonical statistics, are the basis of the vector space and its dimension is the order of the family. Linear exponential families include many common distribution functions, such as multivariate Normal and multinomial distributions. (A linear exponential family in a minimal representation is often called a *regular* exponential family).

A subfamily of linear exponential family is a subset $\mathcal{S}_0$ of $\mathcal{S}$. A subfamily is usually described by a mapping $w \to \eta(w)$ which defines $\mathcal{S}_0$ via $N_0 = \{\eta(w) \mid w \in \Theta\}$ and where $\Theta$ is the domain of $w$. When $\eta$ is a linear mapping of rank $p$, and $\Theta$ is an open set, a new linear exponential family is formed of order $k - p$. In other words, a linear transformation $\eta$ imposes $p$ independent linear constraints on the parameters and these constraints can be used to reparameterize the family with $k - p$ natural parameters. In Sections 3 and 4 we discuss exponential families that are formed by non-linear transformations $\eta$.

### 2.2   Undirected graphical models

In this subsection we discuss the representation of undirected graphical models as linear exponential families.

Let $G$ be an undirected graph such that each vertex $i$ in the vertex set corresponds to a variable $x_i$. We consider three cases. If all $x_i$ are discrete, if all are continuous and their joint density is a multivariate non-singular Gaussian, and if some are continuous and some are discrete with a joint Conditional Gaussian (CG) distribution. An *undirected graphical model w.r.t. G* is the set of probability distribution functions such that all of the saturated independence facts implied by the graph hold; that is $x_i$ and $x_j$ are conditionally independent given the remaining variables whenever nodes $i$ and $j$ are not adjacent in $G$. Since multinomial, multivariate Gaussian and CG distributions over a fixed set of variables belong to a linear exponential family and since saturated independence constraints are linear restrictions when expressed in terms of the natural parameters, undirected graphi-

$\{x_1, \ldots, x_{i-1}\}$, called the *parents set* of $x_i$, and let $u_i = \{x_1, \ldots, x_{i-1}\} \setminus p_i$. Let $x_i^j$, $p_i^j$ and $u_i^j$ be the $j$th value of $x_i$, $p_i$ and $u_i$ with $j \geq 0$. Let $|x_i|$, $|p_i|$ and $|u_i|$ be the domain sizes respectively. The components of $B_{n,m} : \Theta \subseteq R^n \to R^m$ are defined by $\theta_{x_i^a | p_i^b, u_i^c} = \theta_{x_i^a | p_i^b}$, for all $a > 0$, $b \geq 0$, and $c \geq 0$. Note that $n = \sum_i (|x_i| - 1)|p_i|$ and $m = \sum_i (|x_i| - 1)|p_i||u_i| = (\prod_i |x_i|) - 1$. The set $\Theta$ is the cartesian product of $\Theta_{i,j}$ over $i$ and $j$ where $\Theta_{i,j} = \{(\theta_{x_i^1 | p_i^j}, \ldots, \theta_{x_i^{a-1} | p_i^j}) | 0 < \theta_{x_i^l | p_i^j} < 1, \sum_{k > 0} \theta_{x_i^k | p_i^j} < 1\}$. The components of the image of $\Theta$ under $B_{n,m}$ are called the *conditional-space parameters*.

**Theorem 2** *For every Multinomial Bayesian network $B(\Theta, n, m)$ the set $B_{n,m}(\Theta)$ is an n-dimensional manifold in $R^m$.*

**Proof:** Define the components of a function $h$ by $h_{a_i, b_i, c_i}(\theta) = \theta_{x_i^a | p_i^b, u_i^c} - \theta_{x_i^a | p_i^b, u_i^0}$ where $a > 0$, $b \geq 0$ and $c > 0$. Thus, $h$ has $\sum_i (|x_i| - 1)|p_i|(|u_i| - 1) = m - n$ components. Note that $h^{-1}(0) = B_{n,m}(\Theta)$. Also note that $h'$ has the form $[Q_{(m-n) \times n} | I_{m-n}]$ where $I_{m-n}$ is the identity matrix and so $h'$ has full rank. Thus, according to Theorem 1, $B_{n,m}(\Theta)$ is a $n$-dimensional manifold in $R^m$. □

A second definition of a multinomial Bayesian network $\hat{B}$ is obtained by defining $\hat{B}_{n,m}$ with the equations: $w_{a_1, \ldots, a_k} = \prod_{i=1}^k \theta_{x_i^b | p_i^c}$ where $b$ and $c$ are values of $x_i$ and $p_i$ obtained by the projection of $(a_1, \ldots, a_k)$ to the coordinates that correspond to these variables. The mapping $B_{n,m}(\Theta) \to \hat{B}_{n,m}(\Theta)$ is a diffeomorphism for positive $\theta$ values and so the conclusion of Theorem 2 remains valid under this definition as one would expect. The components of the image of $\Theta$ under $\hat{B}_{n,m}$ are called the *joint-space parameters*.

The practical significance of Bayesian networks stems, among other reasons, from the small number of network parameters compared to the number of joint-space parameters. When the number of network parameters is still too large because $|p_i|$ is too large for some $i$'s, additional factorizations are usually introduced. These include decision tree and decision graph models (Friedman and Goldszmidt 1996; Chickering, Meek, and Heckerman, 1997), noisy-or gates, leaky noisy-or gates, max-gates and causal independence models (Pearl, 1988, Henrion, 1987, and Heckerman and Breese, 1996). These models share the following characteristic.

For each variable $x_i$ in the Bayesian network, a subset of $k_i$ states of $p_i$ are designated as *reference states*. The components of $B_{n,m} : \Theta \subset R^n \to R^m$ are defined by $\theta_{x_i^a | p_i^b, u_i^c} = f_i(\theta_{x_i^a | p_i^0}, \ldots, \theta_{x_i^a | p_i^{k_i-1}})$ for all $a > 0$, $b \geq k_i$, and $c \geq 0$ where $f_i$ are smooth functions. We call Bayesian networks defined in this way Bayesian networks with *explicit local constraints*. The number of network parameters is given by $n = \sum_i (|x_i| - 1)k_i$ where $k_i$ is often much smaller than $p_i$.

When the number of reference states is zero, namely each $f_i$ is the constant function, we get a multinomial Bayesian network. In the case of a noisy-or model the reference states are the states where exactly one parent is on and the other parents are off. For leaky noisy-or model the reference states also include the state when all the parents of $x_i$ are off. For decision tree models, the reference states are those which correspond to a path from the root to a leaf in the decision tree; all parents on the path are at a specified state and all those not on the path are at state zero.

**Theorem 3** *For every Bayesian network $B(\Theta, n, m)$ having explicit local constraints the set $B_{n,m}(\Theta)$ is an n-dimensional manifold in $R^m$.*

**Proof:** Suppose the local constraints are given by $f_i$. Define the components of a function $h$ by

$$h_{a_i, b_i, c_i}(\theta) = \theta_{x_i^a | p_i^b, u_i^c} - f_i(\theta_{x_i^a | p_i^0 u_i^0}, \ldots, \theta_{x_i^a | p_i^{k_i-1} u_i^0})$$

where $(a > 0,\ b \geq 0,\ c > 0)$ or $(a > 0,\ b \geq k_i,\ c = 0)$. Note that $h$ has $\sum_i (|x_i| - 1)[|p_i|(|u_i| - 1) + (|p_i| - k_i)] = m - n$ components. The conclusion now follows from Theorem 1. □

Recall that for a multinomial distribution with $u$ states each associated with a positive parameter $w_i$ such that $\sum_i w_i = 1$, the map $\eta_i = \log w_i / w_0$, $i = 1, \ldots, u - 1$ defines a diffeomorphism between the natural parameter space $\eta$ and the multinomial parameters $\{w_i\}_0^{u-1}$. Consequently, due to Theorem 2, we have established the following claim.

**Theorem 4** *Every Bayesian network $B(\Theta, n, m)$ with explicit local constraints is a curved exponential model of dimension $n$.*

We note that the results of Heckerman and Meek (1997), while applied to a different class of models, essentially show that multinomial Bayesian networks are CEFs.

### 3.3  Gaussian graphical models

The parameters of a multivariate non-singular Gaussian distribution can be described in various ways. The most common representation is by the elements of a covariance matrix $\Sigma$ and a vector of means $\mu$. A second representation is by a precision matrix $\Sigma^{-1}$ and $\mu$. These two representations are related by the diffeomorphism $f : \Sigma \to \Sigma^{-1}$. A third representation is constructed as follows. Assign a total order to the $k$ variables. Specify the regression coefficients $b_{i,j}$ of $x_i$ given $x_1, \ldots, x_{i-1}$, and the conditional variance and conditional means of $x_i$ given $x_1, \ldots, x_{i-1}$. The third representation is called the *regression parameterization* and is related to the second representation by a well-known diffeomorphism (e.g., Shachter and Kenley, 1989).

A *Gaussian Bayesian network* is a family of multivariate non-singular Gaussian distributions in which some $b_{ij}$ are set to zero (Shachter and Kenley, 1989). A *Gaussian undirected graphical model* was defined in Section 2.2 to be a family of multivariate non-singular Gaussian distributions in which some of the off-diagonal elements of the precision matrix are set to zero. Both models define a map $B_{n,m} : \Theta \subset R^n \rightarrow R^m$. It follows from Theorem 1 that $B_{n,m}(\Theta)$ is a $n$-dimensional manifold in $R^m$ since the components of $h$ can be defined as projections and so $h'$ has the form $[Q_{(m-n) \times n} | I_{m-n}]$ where $I_{m-n}$ is the identity matrix and $Q$ is a matrix of zeros.

The difference between the two models is that the restrictions formed by setting elements of the precision matrix to zero define linear constraints in the natural parameter space and therefore Gaussian undirected graphical models are also LEFs while the restrictions set by a Gaussian Bayesian network are not linear in the natural parameter space. To demonstrate the latter fact we note that the restriction $b_{31} = 0$ imposed by the Gaussian Bayesian network $x_1 \rightarrow x_2 \leftarrow x_3$ can, in terms of the precision parameters, be written as $t_{1,2} t_{3,3} = t_{1,3} t_{2,3}$ and thus is not linear in the natural parameter space. See Geiger and Heckerman (1994) for the relationships between $t_{i,j}$ and $b_{i,j}$ for this three-node model.

We note that Spirtes, Richardson, and Meek (1997) show that Gaussian MAGs define smooth manifolds. Since Gaussian MAGs are a generalization of Gaussian Bayesian networks, their results also imply that Gaussian Bayesian networks define smooth manifolds.

# 4   Semi-algebraic sets, Implicitization, and Stratified Exponential Families

In this section we provide a definition of semi-algebraic sets and then show that graphical models, whether implicitly or parametrically defined, correspond to semi-algebraic sets. We note that semi-algebraic sets define a union of smooth manifolds rather than a single smooth manifold as in Section 3. We then describe a process called *implicitization* that takes a graphical model defined parametrically and generates both independence and non-independence constraints over the observable variables. We illustrate the implicitization process on several Bayesian networks with hidden variables. Finally, we define stratified exponential families (SEFs), a generalization of curved exponential families, and show that graphical models representing Conditional-Gaussian, Gaussian, or multinomial distributions with or without hidden variables are SEFs.

## 4.1   Semi-algebraic sets

The set of all polynomials in $x_1, \ldots, x_n$ with real coefficients is denoted by $R[x_1, \ldots, x_n]$. Let $q_1, \ldots, q_t$ be polynomials in $R[x_1, \ldots, x_n]$. A *variety* $\mathbf{V}(q_1, \ldots, q_t)$ is the set $\{(x_1, \ldots, x_n) \in R^n | q_i(x_1, \ldots, x_n) = 0 \text{ for all } 1 \leq i \leq t\}$. A variety is also called an *algebraic set*.

A subset $V$ of $R^n$ is called a *semi-algebraic set* if $V = \cup_{i=1}^s \cap_{j=1}^{r_i} \{x \in R^n | P_{i,j}(x) \Leftrightarrow_{ij} 0\}$ were $P_{ij}$ are polynomials in $R[x_1, \ldots, x_n]$ and $\Leftrightarrow_{ij}$ is one of the three comparison operators $\{<, =, >\}$. Loosely speaking, a semi-algebraic set is simply a set that can be described with a finite number of polynomial equalities and inequalities. A variety is clearly a semi-algebraic set.

A map $f : X \rightarrow Y$ where $X \subseteq R^n$ and $Y \subseteq R^m$ are semi-algebraic sets, is called *semi-algebraic* if the graph of $f$ is a semi-algebraic set of $R^{n+m}$. Note that if $f$ is a polynomial map then $f$ is a semi-algebraic map because its graph can be described by $m$ polynomial equalities: $y_j - f_j(x) = 0$, where $1 \leq j \leq m$. A key result about semi-algebraic sets is given by the Tarski-Seidenberg theorem (see, e.g., Benedetti and Risler, 1990).

**Theorem 5 (Tarski-Seidenberg)** *Let* $f : X \rightarrow Y$ *be a semi-algebraic map. Then the image* $f(X) \subseteq Y$ *is a semi-algebraic set.*

Now we examine the connection between varieties and smooth manifolds. To show that a variety is a smooth manifold one could apply Theorem 1 where the components of the function $h$ are the polynomials that define the variety. A point of a variety at which the rank of this Jacobian drops below its maximal rank is called an *algebraic singularity*. We can apply Theorem 1 only if there are no algebraic singularities. Consider, for example, the variety $\mathbf{V}(x^2 - y^2 z^2 + z^3)$ which is plotted in Figure 1. The Jacobian matrix of this variety is given by $(2x, -2yz^2, 3z^2 - 2zy^2)$ and thus every point on the $y$-axis is an algebraic singularity.

This variety is not a smooth manifold because, locally, at each point of the y-axis other than the origin the surface looks like the intersection of two smooth manifolds. To prove that the variety $\mathbf{V}(x^2 - y^2 z^2 + z^3)$ is not a smooth manifold it suffices to observe that as we approach any point on the $y$-axis other than the origin we have two tangent planes where each plane contains a tangent vector that is not spanned by the other tangent plane. One might hope that if there are algebraic singularities in a variety then the surface is not a manifold, however, there are examples of smooth manifolds that have algebraic singularities (e.g., the origin in $\mathbf{V}((x^2 + y^2)(y - x^2))$, Kendig, 1977). In general, to prove that a variety is not a manifold one must examine the particular defining polynomials.

In addition, as our example suggests, if one removes all singular points from a variety, the remaining set is a union of smooth manifolds. This Theorem is due to Whitney (See, e.g., Milnor, 1968).

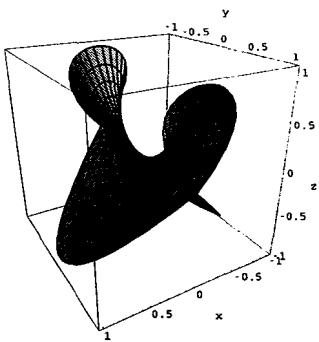

Figure 1: A plot of part of the variety $\mathbf{V}(x^2-y^2z^2+z^3)$.

Another fact about semi-algebraic sets is that they admit a *stratification*. We will define this concept in Section 4.5 but the idea can be illustrated with the variety $\mathbf{V}(x^2 - y^2z^2 + z^3)$. This variety can be described as a union of several 2-dimensional smooth manifolds along with a 1-dimensional smooth manifold— the $y$-axis. These manifolds define a stratification of the variety.

## 4.2    Implicit representations

When the set of distributions associated with a graphical model is given implicitly, as with graphical models without hidden variables, it is straightforward to determine whether or not the model corresponds to a semi-algebraic set. In this section we show that all graphical models defined in previous sections and also other types of graphical models correspond to semi-algebraic sets. In the next subsection we examine graphical models with hidden variables.

The set of distributions associated with a graphical model with structure $g$ is the set of distributions that satisfy all the independence facts entailed by a Markov condition with respect to the structure $g$. For multinomial and Gaussian graphical models an independence fact is expressible as a finite set of polynomial equalities. Combined with the inequalities which state that multinomial parameters are positive, and that variances are positive, respectively, the resulting graphical model corresponds to a semi-algebraic set.

There are several classes of implicitly defined graphical models that can accommodate a combination of discrete and continuous variables using Conditional-Gaussian distributions. Among these models, in addition to all of the models in the previous sections, are AMP chain graphs (Andersson, Madigan, and Perlman, 1996), and reciprocal graphs (Koster, 1997). These graphical models all correspond to semi-algebraic sets because independence facts in CG-distributions are expressible as polynomial equalities.

## 4.3    Parametric representations

In this section we discuss graphical models with hidden variables which are usually defined parametrically. In particular we show that multinomial Bayesian networks with hidden variables correspond to semi-algebraic sets. We note that a similar claim holds for any graphical model representing CG-distributions of which we are aware.

A *Multinomial Bayesian network* $B(\Theta, n, m)$ with hidden variables is a Bayesian network where $\Theta$, $n$, $m$ and $B_{n,m}$ are given as follows. Let $(x_1, \ldots, x_k)$ be an ordered sequence of discrete random variables each having a finite set of values. Partition this set of variables into two disjoint non-empty sets $H$ and $X$. The variables in $H$ are *hidden*. Those in $X$ are *observable*. For each $x_i$ define two disjoint subsets of $\{x_1, \ldots, x_{i-1}\}$. The *observable parents* $p_i \subseteq X$ and the *hidden parents* $h_i \subseteq H$.

The components of $B_{n,m} : \Theta \subseteq R^n \to R^m$ are defined by $w_{a_1,\ldots,a_k} = \sum_d \prod_{i=1}^k \theta_{x_i^{a_i} | p_i^c h_i^d}$ where $a_i$ are not all zero and $b, c, d$ are values of $x_i, p_i, h_i$ obtained by the projection of $(a_1, \ldots, a_k)$ to the coordinates that correspond to $x_i, p_i$ and $h_i$, respectively. As before, the domain $\Theta$ of $B_{n,m}$ is the cartesian product of sets of the form $\{(t_1, \ldots, t_{|x_i|-1}) | 0 < t_a < 1, \sum_a t_a < 1\}$. Note that $n = \sum_{i=1}^k (|x_i| - 1)|p_i||h_i|$ and $m = \prod_{i=1}^k |x_i| - 1$.

The Tarski-Seidenberg theorem guarantees that for a multinomial Bayesian network with hidden variables, $B_{n,m}(\Theta)$ is a semi-algebraic set because it is the image of a semi-algebraic set under a polynomial mapping. Similarly, we note that Gaussian Bayesian network with hidden variables also correspond to semi-algebraic sets due to their parametric definition via a polynomial mapping called the trek-rule (see, e.g., Spirtes et al. 1993). Consequently, the image of these graphical models can be described with a set of polynomial equalities and polynomial inequalities.

## 4.4    Implicitization

As discussed in previous sections, the image of a parametric definition is a semi-algebraic set. The process of taking a parametric definition of a semi-algebraic set and finding a variety (an implicit definition) containing the semi-algebraic set is called implicitization.[1] The implicitization procedure is implemented in several software packages. In our examples we use Mathematica.

Consider the parametric representation of a surface in $R^3$ given by $x = t(u^2 - t^2)$, $y = u$, and $z = u^2 - t^2$.

---

[1]The implicitization procedure is often implemented in software packages by using the Buchberger algorithm for finding Groebner Bases. This algorithm is applied to the polynomial parametric definition. The polynomials in the resulting basis that do not contain any of the parameters are the defining polynomials for the variety (See Cox, Little, O'Shea, 1996).

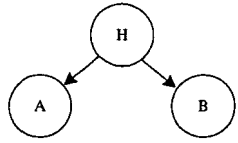

Figure 2: Simple naive Bayes model.

The implicitization procedure applied to this parameterization yields the variety $\mathbf{V}(x^2 - y^2z^2 + z^3)$ discussed in Section 4.1. In this case the surface defined by the variety and the surface defined by the parametric representation are the same, however, this need not be the case. The implicitization procedure is guaranteed to find the smallest variety that contains the image of the polynomial mapping. See Cox, Little, O'Shea (1996) for details and a definition of smallest.

In this section we apply implicitization to various Bayesian networks with hidden variables. It is well known that Bayesian networks with hidden variables entail both independence and non-independence constraints on the distribution of the observed variables. In what follows we illustrate how the implicitization procedure generates such constraints from a parametric definition of a Bayesian network with hidden variables. We start by applying implicitization on a naive Bayesian model generating a previously unknown constraint on the distributions over the observable variables. Then we apply implicitization on two other examples generating two constraints that were previously discovered manually; the tetrad difference constraint and the Verma constraint. In fact, we have applied this technique to several models not discussed in this paper often yielding new constraints on the distributions over the observable variables. The potential use of such constraints for model selection is discussed in Section 5.

Finally we note that the implicitization procedure does not handle inequality constraints. In our examples, we ignore inequality constraints and use only equality constraints. Thus, the resulting implicit representation captures only equality constraints on the joint distributions over the observed variables. We return to this issue when we discuss model selection.

### 4.4.1    Naive Bayes model

In our first example we consider the mapping from the parameters of the Bayesian network to the observable joint parameters (described in Section 3.2) for the naive Bayes model in Figure 2 where $A$ and $B$ are ternary variables and $H$ is binary and hidden. The Mathematica code for implicitization is given in Table 1.

Let $w_{i,j} = P(A = i, B = j)$. The result of implicitization, after algebraic manipulation, are the constraints that $\det(w_{i,j}) = 0$, i.e. the determinant of the joint

```
(* Naive Bayes model with a binary hidden variable and two observed variables each with three values

h = p(H=0)
aij = p(a=j given h=i) except for j=2
bij = p(b=j given h=i) except for j=2
wij = p(a = i, b = j)*)

Eliminate[
w00 == h a00 b00 + (1-h) a10 b10,
w01 == h a00 b01 + (1-h) a10 b11,
w02 == h a00 (1-b00-b01) + (1-h) a10 (1-b10-b11),
w10 == h a01 b00 + (1-h) a11 b10,
w11 == h a01 b01 + (1-h) a11 b11,
w12 == h a01 (1-b00-b01) + (1-h) a11 (1-b10-b11),
w20 == h (1-a00-a01) b00 + (1-h) (1-a10-a11) b10,
w21 == h (1-a00-a01) b01 + (1-h) (1-a10-a11) b11,
w22 == h (1-a00-a01)(1-b00-b01) + (1-h) (1-a10-a11) (1-b10-b11),
a00,a01,a10,a11,b00,b01,b10,b11,h]
```

Table 1: Mathematica code for implicitization.

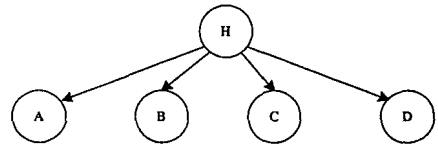

Figure 3: Latent factor model.

parameters is zero, and that $\sum_{i,j} w_{i,j} = 1$.

Unlike the constraint $\det(w_{i,j}) = 0$, the constraints generated by the implicitization procedure for other Bayesian network models did not seem to exhibit such a clear syntactic structure.

### 4.4.2    Tetrad difference constraints

Consider the Gaussian Bayesian network given in Figure 3 where $H$ is not observed. We apply the implicitization procedure to the mapping from the network parameters (i.e., conditional means, regression coefficients, and conditional variances) to the observable parameters (i.e., means and covariance matrix of the multivariate Gaussian distribution). The results are the following two constraints called tetrad difference constraints;

$$cov(A, B)cov(C, D) - cov(A, C)cov(B, D) = 0$$
$$cov(A, B)cov(C, D) - cov(A, D)cov(B, C) = 0.$$

Spirtes et al. (1993) discuss this type of constraints and apply them to the problem of model selection.

### 4.4.3    The P-structure

The final Bayesian network we consider is given in Figure 4 where all variables are binary except the hidden variable $H$ which is ternary. For this example we map the network parameters to the observable conditional-space parameters (Section 3.2) using the ordering $A < B < C < D$.

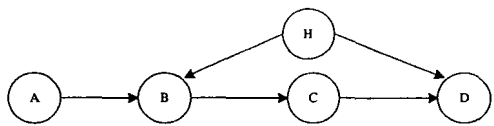

Figure 4: The p-structure.

The results of implicitization on the mapping described above is a set of four constraints; two constraints for the independence between $A$ and $C$ given $B$ and two constraints that we call the Verma constraints. The Verma constraints are discussed in detail in (Spirtes et al., 1993). The Verma constraints are

$$\sum_B (P(B|A = 0)P(D|A = 0, B, C = 0) \\ -P(B|A = 1)P(D|A = 1, B, C = 0)) = 0$$

$$\sum_B (P(B|A = 0)P(D|A = 0, B, C = 1) \\ -P(B|A = 1)P(D|A = 1, B, C = 1)) = 0.$$

We note that these constraints also hold when the edge $b \to c$ is removed from the $P$-structure. The resulting structure is sometimes called the $W$-structure.

#### 4.5   Stratified Exponential Families

A *stratification* of a subset $E$ of $R^m$ is a finite partition $\{A_i\}$ of $E$ such that (1) each $A_i$ (called a *stratum* of $E$) is a smooth $d_i$-dimensional manifold in $R^m$ and (2) if $A_j \cap \overline{A_i} \neq \emptyset$, then $A_j \subseteq \overline{A_i}$ and $d_j < d_i$ (frontier condition) where $\overline{A_i}$ is the closure of $A_i$ in $R^m$. See Akbulut and King (1992) for a more general definition.

A stratification is called *semi-algebraic* if every stratum is semi-algebraic. A *stratified set* is a set that has a stratification. The dimension of a stratified set is $d_1$—the largest dimension of a stratum. A key theorem about semi-algebraic sets is that each semi-algebraic set has a semi-algebraic stratification (Benedetti and Risler, 1990).

We note that if $E$ is a stratified set and $f$ is a diffeomorphism, then $f(E)$ is also a stratified set. This proposition, that stratification is preserved under a diffeomorphism $f$, is proven as follows. Let $\{A_i\}$ be a stratification of $A$. We show that $\{f(A_i)\}$ is a stratification of $f(A)$. Clearly, $\{f(A_i)\}$ is a partition of $f(A)$. Due to Lemma 1, the image of a smooth manifold $A_i$ under a diffeomorphism $f$ is a smooth manifold $f(A_i)$ and so condition (1) of the definition of stratified sets is satisfied. The frontier condition is satisfied because $A_i \subseteq \overline{A_i}$ implies $f(A_i) \subseteq \overline{f(A_i)}$ which, due to continuity of $f$, implies $f(A_i) \subseteq \overline{f(A_i)}$ as needed for satisfying the frontier condition.

We define a *stratified exponential family (SEF)* of dimension $n$ as a subfamily of an exponential family having a natural parameter space $N$ of order $k$ if its parameter space $N_0 \subset N$ is a $n$-dimensional stratified set in $R^k$. Note that SEFs are a proper superset of CEFs.

An examination of all models considered in this paper reveals that $N_0$ defined by each of these models is a stratified set because it is a semi-algebraic set or diffeomorphic to one.

## 5   Asymptotic Model Selection

An important application of the classification of graphical models and the implicition procedure described in the previous sections is model selection. In fact, the work described in this and previous sections is part of an on-going project with David Heckerman of identifying and extending results on asymptotic model selection for directed graphical models with and without hidden variables (e.g. Geiger et al., 1996). In this section we review asymptotic model selection, place our results in this context, and discuss future work.

A Bayesian approach to model selection is to compute the probability that the data is generated by a given model via integration over all possible parameter values with which the model is compatible and to select a model that maximizes this probability. We call this probability the marginal likelihood. Although, in principle, this Bayesian approach is appealing, in practice, it is often impossible to evaluate the integral, even by sampling techniques, when the number of parameters is large. When the dataset consists of many cases, asymptotic results for approximating the marginal likelihood are useful.

Schwarz (1978) considered the problem of evaluating the marginal likelihood when a model is an affine subspace of the natural parameter space of an exponential family. He derived an asymptotic formula for the marginal likelihood, $P(Data|Model) = L(\hat{\theta})N - d/2 \log N + O_p(1)$, where $L$ is the likelihood, $\hat{\theta}$ is the maximum likelihood estimator, $d$ is the dimension of the affine subspace, and $N$ is the sample size. This formula has become known as the Bayesian Information Criteria (BIC). Its plausibility has also been argued using the minimum description length (MDL) principle. We note that Schwarz's original proof applies to the undirected graphical models discussed in Section 2.2 because these models define a linear subspace of the natural parameter space.

In this section we discuss a wider context in which BIC can be justified as an asymptotic Bayesian procedure for selecting models from an exponential family. First, we summarize Haughton's (1988) results for model selection when a model is a smooth manifold (not necessarily affine) of the natural parameter space of an exponential family. Then we discuss how to use constraints and Haughton's results for model selection. Finally, we highlight the difference between CEFs and SEFs and discuss future research directions.

## 5.1  Model selection among CEFs

Haughton (1988) established, under some regularity assumptions, among other results, that BIC is an $O_p(1)$ asymptotic approximation of the marginal likelihood for curved exponential families. The main regularity assumption of her work, and of Schwarz's work, is that the prior distribution expressed in a local coordinate system near the maximum likelihood solution is bounded and bounded away from zero. Other regularity assumptions are used to insure that with sufficient data, a unique model is selected with high probability. When these assumptions are acceptable, Haughton's results on model selection apply to all graphical models discussed in Section 3 since these graphical models have been shown to be curved exponential families. In particular these results on model selection apply to Bayesian networks with several families of local distributions including decision trees and leaky noisy-or distributions for which a closed-form formula for the marginal likelihood is not known.

## 5.2  Model selection using constraints

Graphical models with hidden variables can entail independence and non-independence constraints on the distribution of the observable variables. Since these constraints vary from one model to another they can be used to distinguish between models. Moreover, since these constraints are over the observable variables, their fit to data can be measured directly with statistical tests. In this section we discuss how to use the constraints produced by implicitization for model selection. We concentrate on two examples; the tetrad difference constraints (Section 4.4.2) for which classical statistical techniques have been established, and the constraints implied by the P-structure (Section 4.4.3) for which we adapt BIC.

Gaussian Bayesian networks with hidden variables entail tetrad difference constraints. A classical test of the tetrad difference is provided by a Wishart (1928) significance test. Bollen and Ting (1993) have used these and similar distribution free tests for evaluating the quality of hidden variable models. Spirtes has provided a graphical characterization and a method for calculating tetrad difference constraints from Gaussian Bayesian networks with hidden variables (see, Spirtes et al., 1993). By calculating the set of tetrad difference constraints that are implied by each of a set of competing structures and using the Wishart significance test one can select models from the set of competing structures. A procedure based on this characterization and the Wishart test is implemented in Tetrad II (Scheines, Spirtes, Glymour, and Meek, 1994).

Consider now the situation where we are interested in distinguishing between the P-structure in Figure 4, denoted by $m_1$, and the structure $m_2$ in Figure 5; perhaps we are interested in whether or not $B$ is a direct cause of $D$. Note that the two structures cannot be distinguished by independence facts alone since they have the same entailed independence facts on the observed variables. The model $m_2$ is a curved exponential family (Section 3.2), however, $m_1$ may not be a CEF (it remains an open question to prove that the P-structure is a CEF when $H$ is ternary) and so applying BIC has not yet been justified.

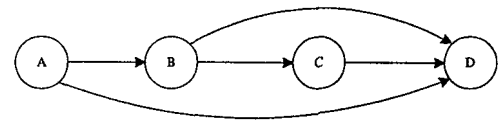

Figure 5: An alternative structure.

As an alternative, we consider the variety defined by the constraints over the observable variables implied by $m_1$. Since the Jacobian matrix of this variety has no algebraic singularities, and due to Theorem 1, we conclude that the set of distributions satisfying these four constraints define a curved exponential family which we denote by $\hat{m}_1$. Note that the set of distributions parameterized by $\hat{m}_1$ is a superset of the distributions parameterized by $m_1$.

Since both $\hat{m}_1$ and $m_2$ are curved exponential families we can use BIC to select between these two models. However, since $m_1 \subseteq \hat{m}_1$, we can only definitively reject $m_1$ in favor of $m_2$. If the selection criterion favors $\hat{m}_1$ there would be some evidence in favor of $m_1$.

## 5.3  Model selection among SEFs

The difficulty in comparing models $m_1$ and $m_2$ in the previous section and the appeal to a super model $\hat{m}_1$ highlights two research questions that need to be addressed. First, whether the graphical models described in Section 4 are CEFs, and second, whether BIC is a valid asymptotic Bayesian rule for selecting models from a stratified exponential family.

We believe that many models described in Section 4 are not CEFs. In particular, with David Heckerman and Henry King, we have examined the Naive Bayesian model with binary observable variables and we believe that this model is not a CEF. Our experience in analyzing whether or not a parametrically defined graphical model is a curved exponential family suggests that the assumption that models describe smooth manifolds is difficult to justify. It is clear, however, that this assumption simplifies many claims because it guarantees the existence of a well-defined tangent space associated with each point in the parameter space.

A question arises, then, whether BIC is valid as an asymptotic procedure for Stratified Exponential Families. The proof of its validity for CEFs in (Haughton, 1988) uses inherent properties of smooth manifolds and so this proof does not extend to SEFs. However, we believe, and are working with David Heckerman to show, that BIC is valid for stratified exponential families for various definitions of stratified sets.

## Acknowledgement

We are grateful to Henry King for many fruitful discussions on the mathematics related to this paper and for referring us to the Tarski-Seidenberg and the stratification theorems, to David Heckerman for many discussions about the relationships between Bayesian networks, CEFs, and model selection, and for helping us to establish that $B_{n,m}(\Theta)$ is a semi-algebraic stratification, and to Steffen Lauritzen for guiding us through the mysteries of exponential families. We have also benefited from conversations with and comments by many other people including Christian Borges, Jennifer Chayes, Mike Freedman, Dominique Haughton, Jim Kajiya, Rob Kass and Paul Vos.

cal models define linear exponential families. We now explicate the three cases.

A *Multinomial undirected graphical model* is a family of probability distributions over a finite set $U$ of variables each having a finite domain such that for some set of pairs of indices $\{(i,j)\}$, $x_i$ and $x_j$ are conditionally independent given $U \setminus \{x_i, x_j\}$. Consider, for example, the graph given by a cycle of size 4 with variables $x_1, \ldots, x_4$ arranged clockwise. Then the independence constraints imposed by this graphical model are that $x_1$ and $x_3$ are conditionally independent given $\{x_2, x_4\}$, and that $x_2$ and $x_4$ are conditionally independent given $\{x_1, x_3\}$. Suppose, for simplicity, that the four random variables are binary (having exactly two states) and denote by $w_i$ the probability of the joint $i$th state of the four binary variables ($1 \le i \le 15$) where $w_0 = 1 - \sum w_i$. Each independence constraint translates to 4 equations of the form $w_i w_j = w_k w_l$. Dividing each equation by $(w_0)^2$ and taking the log, yields 8 linear equations in terms of the natural parameters $\eta_i = \log w_i / w_0$. In general, multinomial graphical models are log-affine models which are LEFs (Lauritzen, 1996, pp 76).

A *Gaussian undirected graphical model* is a family of multivariate non-singular Gaussian distributions in which some of the off-diagonal elements $t_{ij}$ of the precision matrix (the inverse of the covariance matrix) are set to zero. Note that setting $t_{ij}$ to zero is equivalent to requiring that variable $x_i$ and $x_j$ are conditionally independent given the remaining variables. Recalling that a multivariate non-singular Gaussian distribution belongs to a linear exponential family and the fact that setting the off-diagonal elements of the precision matrix to zero is equivalent to placing linear restrictions on the natural parameter space yields the conclusion that Gaussian undirected graphical models are linear exponential families. For details see (Lauritzen, 1996, pp. 124–132).

A *Conditional Gaussian undirected graphical model* is a family of Conditional Gaussian (CG) distributions over a set of discrete and continuous variables defined by a set of saturated independence constraints stating that variables $i$ and $j$ are conditionally independent given the remaining variables. That CG undirected graphical models can be represented as linear exponential families is shown in Lauritzen and Wermuth (1989). See also, Lauritzen (1996, pp. 171–175).

# 3   Curved Exponential Families

A subfamily of a linear exponential family $\mathcal{S}_0 \subseteq \mathcal{S}$ is usually described by a mapping $w \to \eta(w)$ which defines $\mathcal{S}_0$ via $N_0 = \{\eta(w)|w \in \Theta\}$ and where $\Theta$ is an open set. A *curved exponential family* of dimension $n$ is defined to be a subfamily of an exponential family of order $k$ such that $N_0$ is a $n$-dimensional manifold in $R^k$. In this section we provide the definitions of $n$-dimensional manifolds and show that Bayesian net-works correspond to smooth manifolds and are therefore curved exponential families (and not linear exponential models). Conditional-Gaussian Bayesian networks and Conditional-Gaussian chain graphs are also curved exponential models.

Curved exponential families were studied by Efron who explored geometrical interpretation of various statistical measures using these families (e.g., Efron, 1978). A comprehensive treatment of this topic is given by Kass and Voss (1997). We study curved exponential models because the standard asymptotic theory is valid for these models. In particular Haughton's (1988) results on model selection applies to all graphical models discussed in this and the previous section.

## 3.1   Manifolds

A *diffeomorphism* $f : U \subset R^n \to R^m$ is a smooth ($C^\infty$) 1-1 function having a smooth inverse. A subset $M$ of $R^n$ is called a $k$-dimensional *smooth manifold* in $R^n$ if for every point $x \in M$ there exists an open set $U$ in $R^n$ containing $x$ and a diffeomorphism $f : U \cap M \to R^k$. We sometime refer to $M$ as a $k$-dimensional manifold or just as a manifold. Since composition of diffeomorphisms is a diffeomorphism, we get the following proposition.

**Proposition 1** *If $g : A \subset R^n \to B \subset R^n$ is a diffeomorphism, then $M \subseteq A$ is a manifold if and only if $g(M)$ is a manifold and $N \subseteq B$ is a manifold if and only if $g^{-1}(N)$ is a manifold.*

Another way to verify whether a subset of $R^n$ is a manifold is given by the following Theorem (e.g., Spivak, 1965).

**Theorem 1** *Let $A \subset R^m$ be open and let $h : A \to R^{m-n}$ be a smooth function such that $h'(x)$ has rank $m - n$ whenever $h(x) = 0$. Then $h^{-1}(0)$ is a $n$-dimensional manifold in $R^m$.*

Note that the rank of the Jacobian matrix $h'$ in Theorem 1 is $m - n$ if $h$ has the form $h_i(x_1, \ldots, x_m) = x_{n+i} - f_i(x_1, \ldots, x_n)$ for $i = 1, \ldots, m-n$ where $f_i$ are smooth functions because in this case the $(m-n) \times m$ matrix $h'$ factors as $[Q_{(m-n) \times n}|I_{m-n}]$ where $I_{m-n}$ is the identity matrix of size $m-n$.

## 3.2   Multinomial Bayesian networks

A *Bayesian network* $B(\Theta, n, m)$ is a mapping $B_{n,m} : \Theta \subset R^n \to R^m$ where $n$ is the number of *network parameters*, $m$ the number of *observable parameters* and where $\Theta$ and $B_{n,m}$ have specific form depending on the type of the Bayesian network considered.

A *Multinomial Bayesian network* $B(\Theta, n, m)$ is a Bayesian network where $\Theta$, $n$, $m$ and $B_{n,m}$ are given as follows (Pearl, 1988). Let $(x_1, \ldots, x_k)$ be an ordered sequence of discrete random variables each having a finite set of values. Let $p_i$ be a subset of